\PassOptionsToPackage{table,dvipsnames}{xcolor}

\documentclass[10pt,twocolumn,letterpaper]{article}

\usepackage[pagenumbers]{author-kit-CVPR2026-v1-latex/cvpr} 

\definecolor{cvprblue}{rgb}{0.21,0.49,0.74}
\usepackage[pagebackref,breaklinks,colorlinks,allcolors=cvprblue]{hyperref}
\usepackage{hyperref}
\usepackage{url}
\usepackage{graphicx}
\usepackage{multirow}
\usepackage[ruled,vlined,linesnumbered]{algorithm2e}
\SetKwInput{KwInput}{Input}
\SetKwInput{KwOutput}{Output}
\SetKwComment{Comment}{$\triangleright$\ }{}
\SetAlFnt{\footnotesize}
\SetAlCapFnt{\footnotesize}
\SetAlCapNameFnt{\footnotesize}
\SetAlgoNlRelativeSize{-1}
\SetInd{0.4em}{0.8em}
\DontPrintSemicolon
\usepackage{soul}

\usepackage{xcolor}   
\usepackage{colortbl} 
\usepackage{booktabs} 
\usepackage{tikz}
\usetikzlibrary{arrows.meta,shapes.geometric,shapes.symbols,positioning,fit,calc}

\newcommand{\na}{--}

\usepackage{amsmath,amsfonts, mathtools, bm}

\DeclareMathOperator*{\argmin}{arg\,min}

\def\paperID{*****} 
\def\confName{CVPR}
\def\confYear{2026}

\title{From Fewer Samples to Fewer Bits: Reframing Dataset Distillation as Joint Optimization of Precision and Compactness}

\author{
My H. Dinh$^{1}$\thanks{Work completed during a Summer Internship at the AI Lab, InterDigital, Los Altos, CA. Now at Walmart Inc (email: my.dinh@walmart.com).} \quad
Aditya Sant$^{1}$\thanks{Corresponding author.} \quad
Akshay Malhotra$^{1}$ \quad
Keya Patani$^{1}$ \quad
Shahab Hamidi-Rad$^{1}$ \\
\vspace{0.5em}
$^{1}$InterDigital Communications, Inc. \\
\vspace{0.5em}
\texttt{\{first.last\}@interdigital.com}
}
\begin{document}
\maketitle
\begin{abstract}
Dataset Distillation (DD) compresses large datasets into compact synthetic ones that maintain training performance. However, current methods mainly target sample reduction, with limited consideration of data precision and its impact on efficiency.
We propose Quantization-aware Dataset Distillation (QuADD), a unified framework that jointly optimizes dataset compactness and precision under fixed bit budgets. QuADD integrates a differentiable quantization module within the distillation loop, enabling end-to-end co-optimization of synthetic samples and quantization parameters. Guided by the rate–distortion perspective, we empirically analyze how bit allocation between sample count and precision influences learning performance. Our framework supports both uniform and adaptive non-uniform quantization, where the latter learns quantization levels from data to represent information-dense regions better. Experiments on image classification and 3GPP beam management tasks show that QuADD surpasses existing DD and post-quantized baselines in accuracy per bit, establishing a new standard for information-efficient dataset distillation. 
\end{abstract}

\begin{figure*}
    \centering
    \includegraphics[width=0.85\linewidth]{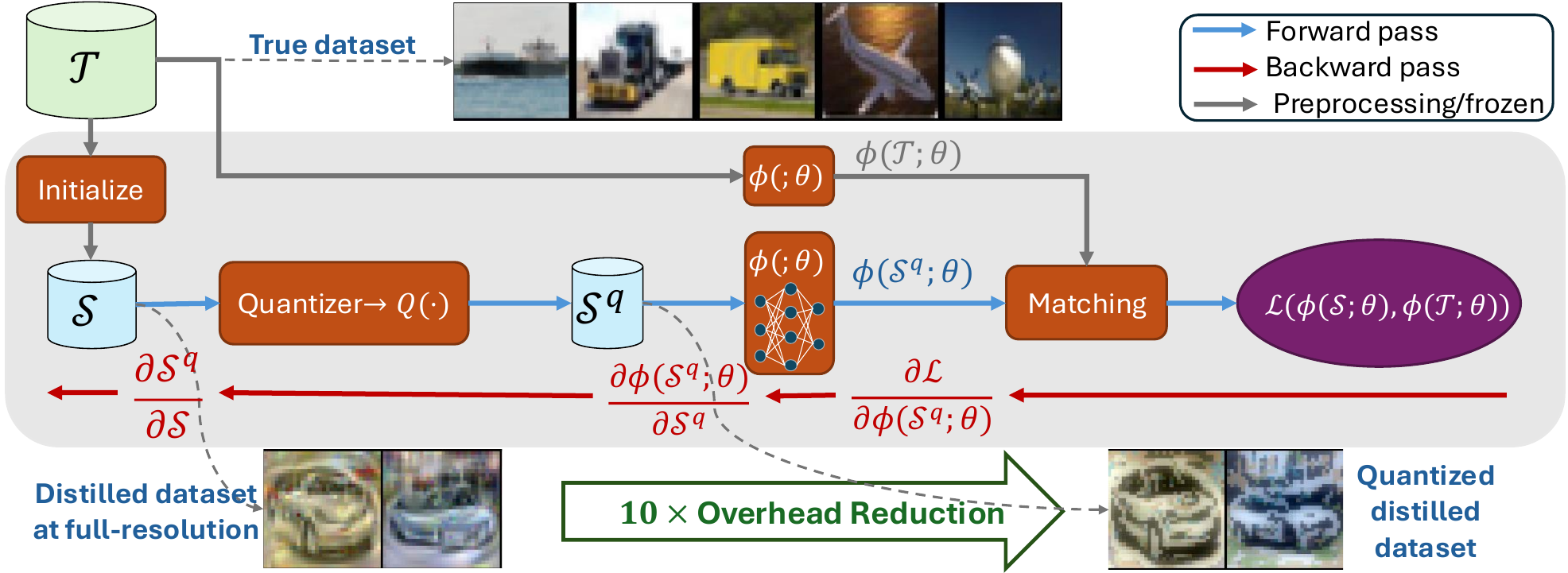}
    \caption{Illustration of the quantized dataset distillation (QuADD) framework, applied to a general DD algorithm.}
    \label{fig:pipeline}
\end{figure*}

\section{Introduction}
In the era of big data, information is continuously generated from diverse sources such as smart devices, sensors, and autonomous systems. 
Sharing and leveraging diverse data is crucial for training robust deep learning models, as sample diversity promotes better generalization. However, the ever-increasing data volume introduces severe computational, storage, and communication bottlenecks, particularly in distributed or resource-constrained environments. 

\noindent 
A natural approach to mitigating these costs is constructing a compact dataset that preserves the essential information of the original data. Traditional methods such as coreset selection~\cite{phillips2016coresetssketches} identify representative subsets of real samples but rely on the assumption that sufficiently informative samples already exist in the dataset.
Dataset Distillation (DD)~\cite{wang2020datasetdistillation} offers a flexible approach by synthesizing a compact set of informative samples that capture the knowledge of the full dataset.
Models trained on the distilled datasets achieve comparable performance to those trained on the original data while requiring far less storage and computation.

\noindent Existing DD methods treat synthetic samples as learnable parameters, optimized to reduce the discrepancy between models trained on real and synthetic data.
These methods improve dataset compactness by reducing training samples.
Recent parameterization-based DD methods extend this idea by representing synthetic data in lower-dimensional spaces that exploit shared structure among samples, yielding further gains in storage efficiency~\cite{wei2023sparse, deng2022rememberpastdistillingdatasets, kimICML22, shin2024}. 

\noindent
Despite these advances, most DD work overlooks a broader notion of data efficiency. 
A dataset is fundamentally an information representation, and its true cost depends not only on the number and dimensionality of samples but on the total bits required to store or transmit it. This becomes critical in practical scenarios like distributed learning, which relies on bandwidth-limited links, and in IoT or edge environments with strict storage constraints. These considerations motivate viewing DD through an information-centric lens, shifting the goal from reducing sample count to minimizing total information content: \textit{from fewer samples to fewer bits}.

\noindent Quantization is a natural solution to control precision. Although widely studied in model and image compression, its integration into DD remains largely unexplored. 
Recent work~\cite{colornips2024} introduced a color-quantization approach that learns a palette-based codebook to merge similar colors, reducing bit depth while preserving salient features like edges and shapes. 
However, this approach is restricted to images, adds training complexity, and scales poorly to other modalities.
These limitations underscore the need for a general framework to jointly optimize dataset compactness and precision under a unified, domain-agnostic objective.

\noindent 
Integrating quantization into DD introduces several challenges. Quantization typically involves two operations: \emph{(i) clipping}, which constrains values to a predefined range by setting any out-of-bound values to the range boundary, and \emph{(ii) projection}, which maps each clipped value to the nearest discrete quantization level. 
Their combination hinders direct optimization via standard gradient descent.
Moreover, quantization introduces information loss that can amplify the reduction already imposed by DD. Thus, an effective quantization scheme must minimize distortion while adapting to the evolving data distribution during distillation. Additionally, it should also remain simple and modality-agnostic to ensure applicability across diverse data types.

\noindent
To address these challenges, we propose Quantization-aware Dataset Distillation (QuADD), shown in Fig.~\ref{fig:pipeline}, a unified framework that integrates a differentiable quantization layer within the distillation loop to \emph{jointly optimize} synthetic data and quantizer parameters end-to-end. The training becomes explicitly aware of precision: synthetic samples adapt to quantization-induced loss, while the quantizer adapts to the distilled data. By coupling quantization with distillation, QuADD shifts the goal from reducing samples to a broader objective of information-efficient learning.

\medskip\noindent{\bf Technical contributions of this work:}
\begin{enumerate}
    \item  \textit{Quantization-aware distillation.} \textbf{QuADD} embeds a differentiable quantizer in the distillation loop, jointly optimizing quantization parameters and synthetic data. Inspired by the rate distortion perspective with explicit bit accounting, we empirically analyze the trade-off between sample count and precision, revealing a “sweet spot’’ between fewer high-precision samples and many low-precision ones.
    \item \textit{Differentiable and adaptive quantization module.}  
    We provide a simple, differentiable quantization layer that supports both uniform and non-uniform techniques.   
    Our primary focus is on the adaptive non-uniform variant, which learns quantization parameters from data to allocate finer bins to information-dense regions, improving task fidelity under the same bit budget.
    \item \textit{Cross-domain validation.}  
    We evaluate our framework across diverse image datasets and a tabular wireless-communication benchmark, where data must be transmitted efficiently over a network and training occurs under resource-constrained conditions, demonstrating that QuADD generalizes beyond visual modalities.
\end{enumerate}


\section{Related Work}\label{Sec:rel_work}

\textbf{Dataset Distillation:} Early DD unrolled training for performance matching~\cite{wang2020datasetdistillation,sucholutsky2021soft,Lei_2024}, later replaced by efficient surrogates such as Gradient~\cite{zhao2021dataset}, Distribution~\cite{zhao2023dataset,zhou2022dataset}, and Trajectory Matching~\cite{cazenavette2022distillation} aligning gradients, features, or optimization paths, respectively.

\noindent \textbf{Redundancy reduction in parameterized DD:}
Beyond reducing sample count, parameterization-based approaches improve storage efficiency by exploiting shared information or structure. IDC~\cite{kimICML22} increases information density via deterministic multi-formation transformations.
HaBa~\cite{liu2022dataset}, RTP~\cite{deng2022rememberpastdistillingdatasets}, and SPEED~\cite{wei2023sparse} share structure across images through bases, memory dictionaries, or sparse epitomic tokens, respectively, while FreD~\cite{shin2024} retains only informative frequency bands.
These methods compact structural dimension but still rely on full-precision data.

\noindent
AutoPalette~\cite{colornips2024} addresses bit-level redundancy through palette-based quantization, encoding images as color index maps and codebooks. However, its color-specific design limits applicability beyond visual data. We extend this direction by introducing a differentiable quantization module that generalizes beyond color-based schemes.
\noindent \textbf{Quantization:}
Quantization reduces numerical precision by mapping continuous values to discrete levels, a core principle in signal compression and modern neural network optimization.
Classical scalar and vector quantizers~\cite{GershoGray1992} balance the rate–distortion trade-off, where using fewer bits lowers storage and transmission costs at the expense of fidelity. Uniform quantization employs fixed step sizes for simplicity but performs poorly on long-tailed distributions. Non-uniform quantization adapts resolution to dense regions, enhancing reconstruction under low-bit constraints.

\noindent
Modern approaches learn quantization mappings from data, including LQ-Net~\cite{zhang2018lqnet}, LSQ~\cite{esser2020learnedstepsizequantization}, QIL~\cite{jung2019qil}, and adaptive log-based methods~\cite{nagel2021whitepaper}.
Beyond static designs, differentiable quantization enables end-to-end learning of discrete mappings via gradient approximations such as the Straight-Through Estimator (STE)~\cite{NIPS2016_d8330f85} or soft relaxations~\cite{gong2019differentiablesoftquantizationbridging,uhlich2020mixedprecisiondnnsneed}.

\section{Problem Set-Up}
Let $\mathcal{T} = \{(x_i, y_i)\}_{i=1}^{N}$ denote the large-scale training dataset and $\mathcal{S} = \{(\tilde{x}_j, \tilde{y}_j)\}_{j=1}^{M}$ 
the compact synthetic dataset, where $M \ll N$.  
DD aims to learn $\mathcal{S}$ such that a model trained on $\mathcal{S}$ performs comparably to one trained on $\mathcal{T}$.  
This can be formulated as a bi-level optimization:
\begin{equation}
    \label{eq:bilevel}
    \begin{aligned}
\mathcal{S}^* &= \argmin_{\mathcal{S}}  \; \;  \mathbb{E}_{\theta \sim \Theta} \; [\, \mathcal{L}(\mathcal{T}; \theta_{\mathcal{S}}^{\star}) \,], \\
\text{where \hspace{2pt}} \; \theta^*_{\mathcal{S}} &= \argmin_{\theta} \;\; \mathcal{L} (\mathcal{S}, \theta)
    \end{aligned}
\end{equation}
\noindent where $\theta$ are the model parameters and $\mathcal{L}(\cdot;\theta)$ is the task loss.
The inner loop optimizes on synthetic data, and the outer loop evaluates on real data.

\noindent
To reduce the heavy computation in eq. \eqref{eq:bilevel}, most DD methods adopt a one-step surrogate objective that matches model responses between real and synthetic data~\cite{zhao2021dataset,zhao2023dataset,cazenavette2022distillation}:
\begin{equation}
\label{eq:dd_loss}
\mathcal{S}^* = 
\argmin_{\mathcal{S}} \,
\mathbb{E}_{\theta \sim \Theta}
\big[\mathcal{L}\big(\phi(\mathcal{T};\theta),\, \phi(\mathcal{S};\theta)\big)\big],
\end{equation}
where $\phi(\cdot;\theta)$ denotes intermediate features, gradients, or trajectories depending on the surrogate choice.

\noindent
Conventional DD improves data efficiency by reducing either the number of synthetic samples $M$ or the data dimension $D$ (e.g., via low-resolution or latent-space parameterizations).  
However, each element of $\mathcal{S}$ is stored at fixed precision—typically 8 or 32-bit, leaving storage efficiency partly unexploited.  
We define the overall bit budget as
\begin{equation}
\label{eq:bit_budget}
\text{Budget} = M \times D \times b,
\end{equation}
where $b$ is the bit precision per data element.

\noindent
Optimizing $M$ or $D$ alone thus neglects precision as a controllable degree of freedom. 
Post-quantization of distilled data is a simple workaround, but it often degrades accuracy since synthetic samples are not optimized for low precision.  

\noindent
To address this, we integrate a differentiable quantizer directly into the DD pipeline, enabling the synthetic data and quantization parameters to co-adapt during training.
This \emph{quantization-aware distillation} produces datasets that remain informative even under tight bit budgets.

\noindent
Finally, this formulation can be viewed through a \textit{rate–distortion} lens, where the bit budget is the “rate” and the performance drop the “distortion.” Although QuADD does not explicitly optimize this function, we empirically study how different $(M, b)$ combinations affect performance under fixed storage budgets.

\section{Quantization-Aware Dataset Distillation}
This section details the Quantization-Aware Dataset Distillation (QuADD) framework. 
We first introduce the training objective and gradient flow through the quantizer, then describe the differentiable quantization framework for both uniform and adaptive non-uniform settings.

\subsection{Training Objectives}
\label{sec:quant_loss}
To incorporate quantization into DD loop, QuADD augments the standard DD loop with a differentiable quantizer $Q(\cdot)$. Given synthetic samples $\tilde{x}_j\!\in\!\mathcal{S}$, the quantized representations are
\begin{equation}
\tilde{x}_j^{q} = Q(\tilde{x_j}), \qquad 
\mathcal{S}^{q} = \{(\tilde{x}^{q}_j, \tilde{y}_j)\}_{j=1}^M.
\end{equation}
\noindent
Instead of matching real and synthetic data in full precision, QuADD enforces performance equivalence between real data $\mathcal{T}$ and quantized synthetic data $\mathcal{S}^q$ as
\begin{equation}
\label{eq:dd_quantized_loss}
\mathcal{S}^* = \argmin_{\mathcal{S}}\; \mathbb{E}_{\theta \sim \Theta} \; 
\big[ \mathcal{L}\big(\phi(\mathcal{T};\theta),\, \phi(\mathcal{S}^q;\theta)\big)\big],
\end{equation}
where the bit precision $b$ of $Q(\cdot)$ controls the storage budget.
%

\medskip
\noindent \textbf{Gradient propagation:}
Quantization introduces nondifferentiable operations (clipping and rounding), requiring special handling for backpropagation. 
Loss gradients with respect to the synthetic data are computed via the chain rule:
\begin{equation}\label{eq:grad_chain}
\frac{\partial \mathcal{L}}{\partial \mathcal{S}}
= \frac{\partial \mathcal{L}}{\partial \phi}
  \frac{\partial \phi}{\partial \mathcal{S}^q}
  \frac{\partial \mathcal{S}^q}{\partial \mathcal{S}},
\end{equation}
where the last term $\partial \mathcal{S}^q/\partial \mathcal{S}$ involves differentiation through the quantizer and is approximated using straight-through or smooth surrogates described next.

\subsection{Differentiable Quantization Layer}
\label{sec:quant_layer}
The quantization layer transforms full-precision synthetic data into quantized data while remaining differentiable, enabling end-to-end optimization with the DD objective.

\noindent Quantization is composed of clipping and projection, as
\begin{equation}
\label{eq:quantizer}
Q(x;\alpha,b)
= \Pi_{Q(\alpha,b)}\;
 \lfloor x;\alpha\rceil
\end{equation}

\noindent
where $\alpha$ is the clipping threshold, $\lfloor \cdot, \alpha \rceil$ clips inputs to $[-\alpha, \alpha]$, and $Q(\alpha,b)$ is the $b$-bit codebook.
$\Pi_{Q(\alpha,b)}(\cdot)$ projects each value to its nearest level. The $\mathrm{round}(\cdot)$ operation within the projection is non-differentiable, thus requires special handling to enable gradient training.

\medskip
\noindent \textbf{Forward pass:}
We explored two different forward modes of the $\mathrm{round}$ operator:

\begin{itemize}[leftmargin=10pt, topsep=0pt]
    \item Hard rounding: $\mathrm{round}(\cdot)$ to the nearest codebook value; applicable to both uniform and non-uniform quantizers.
    \item Soft rounding: a continuous relaxation that approximates rounding \cite{mentzer2024finite, parker2025scaling}; used only for the uniform case.
\end{itemize}

\medskip
\noindent \textbf{Backward pass:} During backpropagation, gradients are estimated as:

\begin{itemize}[leftmargin=10pt, itemsep=4pt]
    \item Straight-through estimator (STE): for hard rounding, treat $Q$ as identity within the clipping range:
    \[
    \frac{\partial x^q}{\partial x} \approx \mathbf{1}(|x|\le \alpha).
    \]
    \item Analytic surrogate soft quantizers use the derivative of the smooth approximation of rounding (see Appendix \ref{sec:uniform}).
\end{itemize}

\subsubsection{Baseline: Uniform quantizer}
\noindent When $Q(\alpha,b)$ is uniformly spaced, this reduces to the standard uniform quantization scheme, widely used in model compression and image coding, given as:
\begin{equation}
  Q^{u}(\alpha,b)
    = \alpha \!\times\!
    \left\{
    -1,\;
    \frac{\pm1}{2^{b-1}-1},\;
    \ldots,\;
    1
    \right\}.  
\end{equation}
\noindent
To reduce gradient discontinuities near the clipping boundary, \emph{companding functions} (e.g., $\tanh$, sigmoid, or CDF-based) offer smooth alternatives to hard clipping, with $\alpha$ fixed by the chosen function (e.g., $\alpha=1$ for $\tanh$). This uniform quantizer serves as the baseline for QuADD.

\subsubsection{Adaptive Non-uniform Quantizer}
\label{Sec:adaptive_quant}

\noindent
The adaptive quantizer adopts the \textbf{Additive Powers-of-Two (APoT)} scheme~\cite{Li2020Additive}, which expresses each quantized value as a sum of scaled powers of two:
\begin{equation}
   Q^{a}(\alpha,b)
    = \gamma \;\times\;
    \Big\{
    \sum_{i=0}^{n-1} p_i \;\Big|\;
    p_i \in 
    \Big\{
    0, 
    \tfrac{1}{2^i},
    \ldots,
    \tfrac{1}{2^{i+(2^k -2)n}}
    \Big\}
    \Big\} 
\end{equation}

\noindent
where $\gamma$ scales the dynamic range to $[-\alpha,\alpha]$, $k$ is the \emph{base bit-width} for each additive term, and $n = b/k$ denotes the number of additive bases.  
This representation achieves non-uniform level spacing, allocating more quantization density to dense regions of the data distribution.

\medskip
\noindent \textbf{Forward pass:}
Here, the forward pass uses hard rounding to project values onto additive combinations of PoT bases.

\medskip
\noindent \textbf{Backward pass:}
The clipping threshold $\alpha$ is learnable and updated automatically through back-propagation, allowing the quantizer to adapt to the evolving data distribution. Because the standard STE updates $\alpha$ only from out-of-range values, we adopt the Reparameterized Clipping Function (RCF)~\cite{Li2020Additive}, which rescales inputs to a normalized range before projection and restores their scale afterward.  
This allows $\alpha$ to receive gradients from all samples:
\[
\frac{\partial x^q}{\partial \alpha} =
\begin{cases}
\operatorname{sign}(x), & \text{if } |x| > \alpha, \\[6pt]
\Pi_{Q(1,b)}\!\left(\frac{x}{\alpha}\right) - \frac{x}{\alpha}, & \text{if } |x| \le \alpha.
\end{cases}
\]

\noindent
\textbf{Remarks.}
While APoT was originally designed for model quantization, its adaptation to dataset distillation requires careful initialization and normalization for stable convergence (see Appendix \ref{sec:apot_setting}).
In QuADD, it is computationally efficient, with only one learnable parameter ($\alpha$), and achieves higher fidelity than uniform quantization, making it the default choice in all experiments.

\subsection{Algorithm Implementation for QuADD}
\label{sec:init}
\noindent \textbf{Quantization-guided initialization:}
QuADD initializes synthetic data following the quantization-guided selection strategy of~\cite{colornips2024}, adapted from color to scalar quantization.
The real dataset $\mathcal{T}$ is first quantized using a uniform quantizer to obtain $\mathcal{T}^q = Q(\mathcal{T})$. Let $A \subset \mathcal{T}^Q$ and $C = \mathcal{T}^Q \setminus A$ the remaining candidates. 
Representative samples are iteratively chosen to maximize the condition gain defined by a generalized graph-cut criterion:
\begin{equation}\label{eq_init_eq_2}
    G^*(A|C) = \sum_{i \in C}\sum_{j \in A}\text{Sim}(i,j)
- \!\!\sum_{j_1,j_2 \in A}\!\!\text{Sim}(j_1,j_2)
\end{equation}
\noindent
It measures how much new information a candidate image adds to the selected set.
\noindent Sample similarity $\text{Sim}(i,j)$
 is defined as the cosine similarity of last-layer gradients.
\begin{equation}\label{eq_init_eq_3}
\text{Sim}(i,j)
= \cos \big(
\nabla_\theta L_{\text{CE}} \; (x^q_i,\theta), \,
\nabla_\theta L_{\text{CE}}(x_j^q, \theta)
\big)
\end{equation}
and $\nabla_\theta L_{\text{CE}}$ denotes the gradient of the cross-entropy loss with respect to model parameters.
\begin{algorithm}[t]
\caption{General Quantization-Aware Dataset Distillation}
\label{alg:quadd_general}
\KwInput{Dataset $\mathcal{T}$, quantizer $Q(\cdot;\alpha,b)$, DD framework $\phi(\cdot)$ and loss $\mathcal{L}_{\text{distill}}(\cdot,\cdot)$}
\KwOutput{Quantized distilled dataset $\mathcal{S}^{q^{*}} = Q^*(\mathcal{S}^*)$}
Initialize $\mathcal{S}^{q}$ using \eqref{eq_init_eq_2}-\eqref{eq_init_eq_3} and quantizer parameters $(\alpha,b)$\;
\For{each DD iteration}{
    Sample mini-batch $\mathcal{B}_{\mathcal{T}}\!\sim\!\mathcal{T}$\;
    Quantize synthetic data: $\mathcal{S}^q = Q(\mathcal{S})$\;
    Compute loss $\mathcal{L}_{\text{distill}}\!\left(\phi(\mathcal{\mathcal{B}_{\mathcal{T}}}),\phi(\mathcal{S}^q)\right)$\;
    Backpropagate through both $\mathcal{S}$ and $Q$ using \eqref{eq:grad_chain} \;
    Update $\mathcal{S}$ and $Q$ via gradient descent\;
}
\end{algorithm}
\medskip

\noindent \textbf{QuADD implementation:} Algorithm \ref{alg:quadd_general} outlines the end-to-end procedure for QuADD.
At each iteration, real and synthetic batches are processed through the differentiable quantizer, and the DD loss is computed on quantized data.
Gradients flow through the synthetic samples and quantizer, enabling their co-optimization.

\noindent By varying sample count and bit precision, QuADD empirically traces the rate–distortion performance (see Sec. \ref{Sec:rate_distortion}). The general quantizer implementation integrates with different DD objectives (see Sec. \ref{Sec:exp_ablation}), as well as generalizes across data domains (see Sec. \ref{Sec:exp_3gpp}).

\section{Experiments}

\subsection{Experimental Setup}
We evaluate QuADD across both vision and tabular modalities to demonstrate its generality under fixed storage budgets, using CIFAR-10/100~\cite{krizhevsky2009learning}, ImageNette~\cite{howard2019imagenette}, and a 3GPP beam management dataset (see Appendix \ref{sec:beam_meta} for details).
All models are trained under a fixed storage constraint as defined in eq. \eqref{eq:bit_budget}.
Unless stated otherwise, full-precision baselines use 32-bit data, while QuADD employs 9-bit precision for image data (3~bits per RGB channel) and 8-bit for tabular data to ensure comparable bit budgets.

\subsubsection{Benchmarks and Baselines}
\noindent \textbf{Image classification:}
We compare QuADD against:  
(i) full-precision dataset distillation methods trained without quantization, including DD~\cite{wang2020datasetdistillation}, DM~\cite{zhao2023dataset}, and TM~\cite{cazenavette2022distillation};  
(ii) AutoPalette~\cite{colornips2024}, a color-quantization approach that reduces bit depth~$b$; and  
(iii) FreD~\cite{shin2024}, a parameterization-based DD method that compresses feature dimension~$D$.

\medskip
\noindent \textbf{3GPP beam management:}
For the tabular domain, we evaluate QuADD on a 3GPP beam management dataset, where the goal is to predict the optimal beam index given a \textit{subset} of beam-level measurements.  
We compare QuADD against:  
(i) the full-precision original dataset,  
(ii) a distilled dataset trained by DATM~\cite{guo2024lossless} without quantization, and  
(iii) a coreset-selection baseline using a conditional information-gain objective (see Sec.~\ref{sec:init}).
\begin{figure}[t]
\centering
\includegraphics[width=0.8\columnwidth]{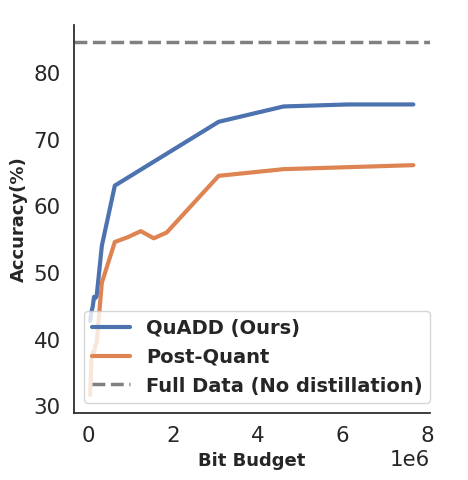}
\caption{Rate–distortion curve.
Each point represents the best achievable accuracy under a fixed transmission budget (in bits).}
\label{fig:rate_distortion}
\end{figure}
\subsubsection{Implementation Details}\label{Sec:implementation}
For image benchmarks, we use 1, 10, and 50 images per class (IPC) following standard DD settings.
ConvNetD3 and ConvNetD5 serve as student and evaluation networks for CIFAR and ImageNette, respectively.
We adopt trajectory-matching ~\cite{cazenavette2022distillation} from DATM~\cite{guo2024lossless} without soft-label initialization.
Results are averaged over five runs with different seeds, with all training details in Appendix \ref{sec:hyperparam}. 

\subsection{Results on Image Benchmarks}
\subsubsection{Rate-distortion Analysis}\label{Sec:rate_distortion}
Figure~\ref{fig:rate_distortion} illustrates the rate–distortion relationship between total transmitted bits and test accuracy.
The bit rate follows eq. \eqref{eq:bit_budget}, combining both the number of synthetic samples and their quantization precision, and each point reflects the best-performing $(M, D)$ configuration under a fixed budget.

\noindent
Higher bit budgets naturally yield better accuracy, but QuADD delivers substantially stronger rate–distortion efficiency than post-training quantization. Across multiple budgets, QuADD improves accuracy by at least 10\%, showing that quantization-aware distillation enables synthetic data to adapt to precision limits and preserve more task-relevant information.

\begin{figure}[t]
\centering
\includegraphics[width=0.8\columnwidth]{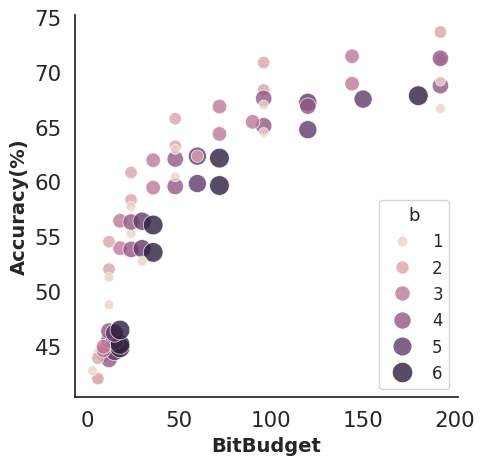}
\caption{QuADD performance on CIFAR-10 under varying $(M,b)$.
Lighter \& smaller markers indicate lower-bit precision.
The storage budget scales as $b \times 3 \times D \times M$ (constants omitted).}
\label{fig:bit_acc}
\end{figure}

\noindent
Figure \ref{fig:bit_acc} illustrates the performance of QuADD on CIFAR-10 as a function of the number of samples $M$ and bit precision $b$. The x-axis represents the total bit budget, and points that align vertically correspond to different 
$(M,b)$ configurations that yield the same total number of transmitted bits. For each such budget, multiple feasible pairs $(M,b)$ exist—for example, a configuration with more samples at lower precision or fewer samples at higher precision.
Across these vertical groups, we observe a consistent trend:
For a fixed bit budget, lower-bit quantization combined with more samples often achieves higher accuracy than using fewer high-precision samples. A clear pattern emerges: for nearly every budget, using lower-bit precision to allow more synthetic samples yields higher accuracy than allocating the budget to high-precision samples. In contrast, using the budget for higher precision limits how many samples can be transmitted, which often yields less effective use of the available bits. Notably, most of the best results come from 2–3 bits per sub-pixel, showing that aggressively reducing precision not only preserves accuracy but often improves the rate–distortion trade-off. This highlights an advantage of QuADD: precision reduction is an effective way to maximize the informational value of transmitted synthetic data

\subsubsection{Main Comparison under Fixed Storage Budget}

\begin{table*}[t]
\centering
\caption{
Comparison of dataset distillation on \textbf{CIFAR-10}, \textbf{CIFAR-100}, and \textbf{Imagenette} under different IPC budgets.
Best result for Parameterized DD in each column is \textbf{bolded}.
For AutoPalette, we quantize images to \textbf{7-bit}; for QuADD, to \textbf{9-bit (3 bits/channel)}; for FreD, we use their best reported hyperparameters. Higher value indicates better performance.}
\label{tab:main-acc}
\renewcommand{\arraystretch}{1.12}
\setlength{\tabcolsep}{5pt}
\begin{tabular}{l l | ccc | ccc | c}
\toprule
& & \multicolumn{3}{c|}{\textbf{CIFAR-10}} & \multicolumn{3}{c|}{\textbf{CIFAR-100}} & \textbf{Imagenette} \\
\cmidrule(lr){3-5}\cmidrule(lr){6-8}
\multicolumn{2}{c|}{}
& IPC 1 & IPC 10 & IPC 50
& IPC 1 & IPC 10 & IPC 50
& IPC 10 \\
\midrule
\multicolumn{2}{l|}{Total storage}
& 983K & 9.8M & 49.1M
& 9.8M & 98.3M & 491M
& 157.2M \\
\midrule
\multirow{3}{*}{Coreset}
& Random         & 14.4$\pm$0.2 & 26.0$\pm$1.2 & 43.4$\pm$1.0 &  4.2$\pm$0.3 & 14.6$\pm$0.5 & 30.0$\pm$0.4 & \na \\
& Herding \cite{10.1145/1553374.1553517} & 21.5$\pm$1.3 & 31.6$\pm$0.7 & 40.4$\pm$0.6 &  8.4$\pm$0.3 & 17.3$\pm$0.3 & 33.7$\pm$0.5 & \na \\
& K-Center \cite{sener2018activelearningconvolutionalneural}  & 23.3$\pm$0.9 & 36.4$\pm$0.6 & 48.7$\pm$0.3 &  8.6$\pm$0.3 & 20.7$\pm$0.2 & 33.6$\pm$0.4 & \na \\
\midrule
\multirow{5}{*}{\shortstack{Unquantized\\DD}}
& DD \cite{wang2020datasetdistillation}       & \na          & 36.8$\pm$1.2 & \na          & \na          & \na          & \na          & \na \\
& DM \cite{zhao2021dataset}        & 26.0$\pm$0.8 & 48.9$\pm$0.6 & 63.0$\pm$0.4 & 11.4$\pm$0.3 & 29.7$\pm$0.3 & 43.6$\pm$0.4 & \na \\
& DC \cite{zhao2023dataset}        & 28.3$\pm$0.5 & 44.9$\pm$0.5 & 53.9$\pm$0.5 & 12.8$\pm$0.3 & 25.2$\pm$0.3 & \na          & \na \\
& TM \cite{cazenavette2022distillation}         & 45.8$\pm$0.8 & 65.1$\pm$0.7 & 71.6$\pm$0.2 & 24.3$\pm$0.3 & 40.1$\pm$0.4 & 47.7$\pm$0.2 & 67.3$\pm$0.3 \\
& DATM \cite{guo2024lossless}      & 46.1$\pm$0.5 & 65.7$\pm$0.2 & 75.1$\pm$0.6 & 27.1$\pm$0.2 & 47.3$\pm$0.4 & 53.9$\pm$0.2 & 67.8$\pm$0.3 \\
\midrule
\multirow{3}{*}{\shortstack{Parameterized  \\DD}}
& FreD \cite{shin2024}        
  & 43.1$\pm$1.5 & 57.3$\pm$1.1 & 69.0$\pm$0.8 
  & 15.7$\pm$0.8 & 34.9$\pm$0.6 & 41.5$\pm$0.5 
  & 66.2$\pm$0.3 \\
& AutoPalette \cite{colornips2024}
  & 42.1$\pm$0.8 & 63.5$\pm$0.4 & 73.9$\pm$0.3 
  & 25.8$\pm$0.3 & 45.6$\pm$0.4 & \textbf{52.9$\pm$0.3} 
  & 63.0$\pm$0.7 \\
\rowcolor{gray!14}
& QuADD (Ours)
  & \textbf{44.9$\pm$0.4} & \textbf{65.4$\pm$0.2} & \textbf{74.8$\pm$0.2} 
  & \textbf{26.9$\pm$0.3} & \textbf{46.2$\pm$0.1} & 52.4$\pm$0.2 
  & \textbf{67.0$\pm$0.4} \\
\midrule
\multicolumn{2}{l|}{Entire dataset}
& \multicolumn{3}{c|}{84.8$\pm$0.1}
& \multicolumn{3}{c|}{56.2$\pm$0.3}
& 87.4$\pm$0.3 \\
\midrule
\midrule
\multicolumn{9}{l}{Overhead reduction (relative to original dataset)} \\[-2pt]
\cmidrule{1-9}
& FreD         
  & $\mathbf{12}\times$ & $9.6\times$ & $6.1\times$ 
  & $\mathbf{12}\times$ & $3.8\times$ & $3.8\times$ 
  & $12\times$ \\
& AutoPalette  
  & $9.6\times$ & $9.6\times$ & $9.6\times$
  & $9.6\times$ & $9.6\times$ & $9.6\times$
  & $\mathbf{13.3}\times$ \\
& QuADD (Ours)
  & $10.6\times$ & $\mathbf{10.6}\times$ & $\mathbf{10.6}\times$
  & $10.6\times$ & $\mathbf{10.6}\times$ & $\mathbf{10.6}\times$
  & $10.6\times$ \\
\bottomrule
\end{tabular}
\end{table*}

\begin{figure}
\centering
\includegraphics[width=1.0\columnwidth]{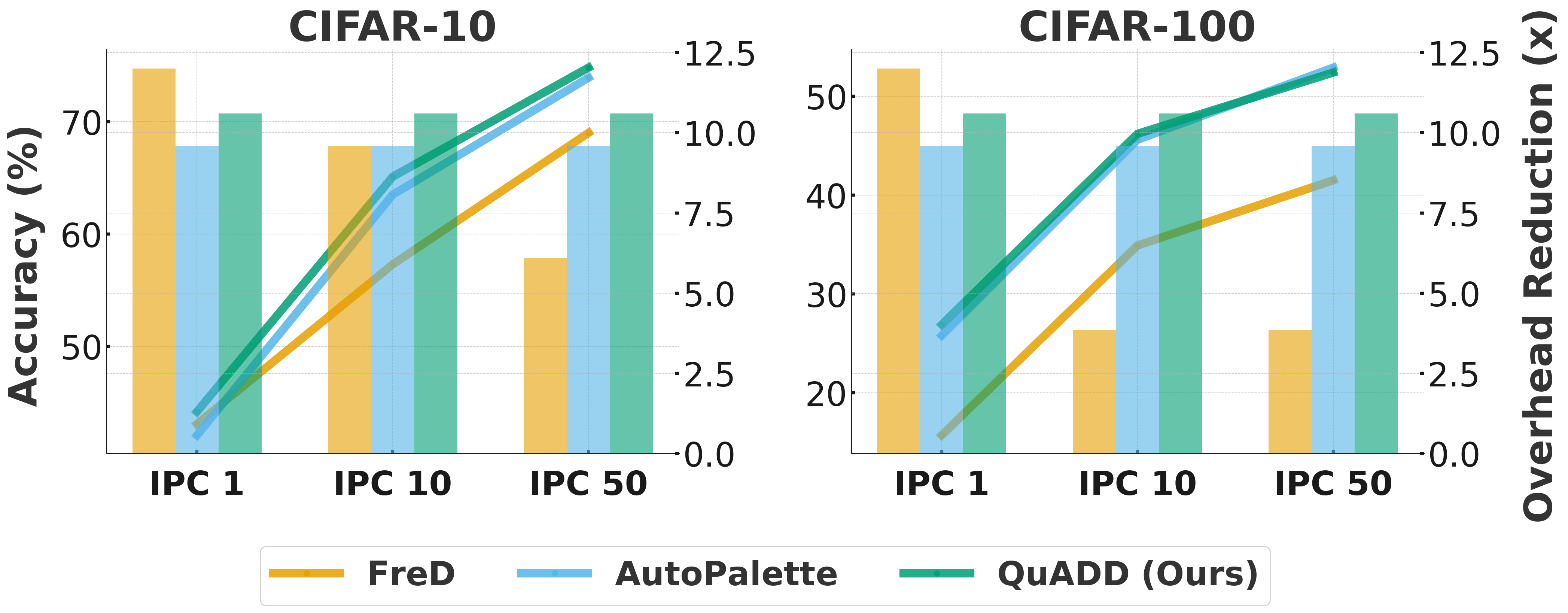}
\caption{Accuracy and overhead reduction on CIFAR-10/100.
Lines show test accuracy (left $y$-axis); bars show compression ratio (right $y$-axis) relative to the unquantized baseline.}
\label{fig:benchmark_main}
\end{figure}

Table~\ref{tab:main-acc} summarizes results across standard vision benchmarks. 
Coreset and unquantized DD methods, as well as FreD~\cite{shin2024}, output 32-bit full-precision values, while AutoPalette~\cite{colornips2024} and our QuADD compress data into lower-bit representations under equivalent configurations. 
To ensure fair comparison, all distillation methods use the same number of synthetic samples per class, i.e., the same IPC/SPC, without increasing sample count to offset quantization—as done in certain storage-budget analyses by AutoPalette and FreD.  
The ``Overhead Reduction'' section denotes the compression ratio relative to the unquantized or coreset baseline, which serves as the full-precision reference.  
We adjust quantization resolution such that the effective storage budgets are closely matched across methods, allowing accuracy to be compared under comparable conditions.

\noindent
QuADD consistently achieves nearly the same accuracy—typically within $1\%$—as the full-precision distilled baseline while using less than one-tenth of the storage. 
For example, on CIFAR-10 with IPC=10, QuADD attains 65.1\% accuracy at $10.6\times$ lower storage, closely matching the 65.5\% accuracy of the unquantized baseline. 
Similar trends hold for CIFAR-100 and Imagenette, confirming that quantization-aware training effectively compensates for bit-level precision loss and preserves task-relevant information even under low-resolution encoding.

\noindent
Figure~\ref{fig:benchmark_main} visualizes these results on CIFAR-10 and CIFAR-100. 
Line plots represent test accuracy (left $y$-axis), while bar plots show overhead reduction ratios (right $y$-axis). 
Across all IPC settings, QuADD consistently achieves higher accuracy under comparable or greater compression—improving average accuracy by $1.5$--$3.0$\% at roughly $10\times$ smaller storage. 
Across various datasets, QuADD outperforms both AutoPalette and FreD, barring the IPC=1 instance, which highlights its enhanced storage efficiency at higher capacities. 
Moreover, while AutoPalette and FreD are primarily tailored for image data, the QuADD framework is modality-agnostic; this is further demonstrated by the subsequent results. in Sec. \ref{Sec:exp_3gpp}.


\subsubsection{Training Efficiency}
\begin{figure}
    \centering
    \includegraphics[width=0.83\linewidth]{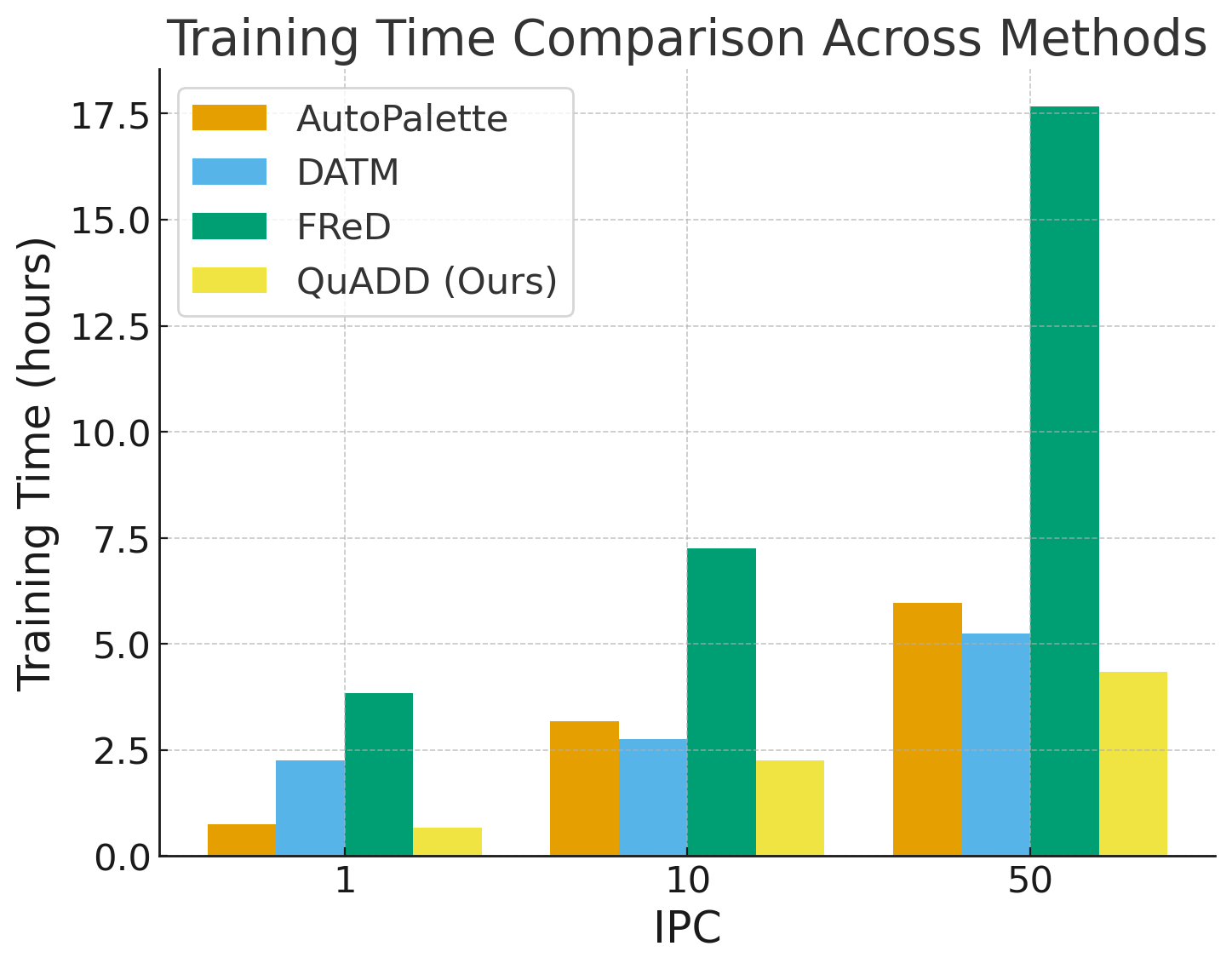}
    \caption{Training time on CIFAR-10 for DD baselines (DATM, FreD, AutoPalette, QuADD) at IPC = 1, 10, 50. Lower is better.}
    \label{fig:train_time}
\end{figure}

Figure \ref{fig:train_time} demonstrates that QuADD maintains competitive or superior training efficiency compared to existing distillation methods. Across all IPCs, QuADD either matches or outperforms the unquantized baseline (DATM), achieving up to 25–30\% reduction in training time at medium and high IPC. In contrast, AutoPalette is slower at all scales, reflecting the additional overhead of its color-space optimization. FreD exhibits the highest cost, with training time growing rapidly with IPC, highlighting poor scalability. Overall, this confirms that QuADD’s quantization layer is simple, lightweight, and adds no extra computation.

\subsubsection{Ablation and Analysis}\label{Sec:exp_ablation}
%
\noindent \textbf{1) Cross architecture performance:}
We evaluate whether QuADD’s distilled datasets transfer beyond the ConvNetD3 teacher by training AlexNet, VGG, and ResNet on the same synthetic data. As shown in Table~\ref{tab:cross_arch}, QuADD maintains strong performance across backbones and IPC settings, outperforming DATM and AutoPalette and remaining competitive with FreD (e.g., 51.8\% on ResNet at IPC=10 vs. 49.0\% for DATM and 50.1\% for AutoPalette).

\noindent
The higher accuracy gains observed by FreD are attributed to the alignment of frequency-domain representations with architectural biases and their capacity to preserve high-frequency details. However, these methods involve certain trade-offs: frequency-domain training is slower, optimization is less stable, and DCT-based parameterizations do not easily generalize to all data types.



\medskip
\noindent \textbf{2) Cross distillation performance:}
To evaluate the generality of the QuADD framework, we implement it on additional DD backbones, as shown in Table \ref{tab:quadd_cross_dd}. Under TM, QuADD maintains high accuracy across all IPC budgets, remaining within ~1–2 points of the full-precision baseline and surpassing FreD and AutoPalette at IPC1 and IPC10. In the more challenging DM setting, QuADD significantly improves over AutoPalette at low IPC and remains competitive with FreD, narrowing the gap as IPC increases



\begin{table*}[t]
\centering
\renewcommand{\arraystretch}{1.2}
\setlength{\tabcolsep}{4pt} 
\resizebox{\textwidth}{!}{%
\begin{tabular}{l|ccc|ccc|ccc}
\toprule
\multirow{2}{*}{Method} 
  & \multicolumn{3}{c|}{AlexNet} 
  & \multicolumn{3}{c|}{VGG} 
  & \multicolumn{3}{c}{ResNet} \\
\cmidrule(lr){2-4} \cmidrule(lr){5-7} \cmidrule(lr){8-10}
 & IPC1 & IPC10 & IPC50 & IPC1 & IPC10 & IPC50 & IPC1 & IPC10 & IPC50 \\
\midrule
Unquantized Baseline (DATM) & 15.6$\pm$1.0 & 25.5$\pm$1.5 & 54.1$\pm$1.3 & 29.8$\pm$0.9 & 39.5$\pm$1.0 & 63.1$\pm$0.3 & 30.1$\pm$1.0 & 49.0$\pm$1.5 & 65.2$\pm$1.0 \\
\midrule
FreD        & \textbf{26.8$\pm$1.2} & \textbf{35.3$\pm$1.5} & \textbf{56.9$\pm$1.3} & \textbf{34.7$\pm$0.8} & \textbf{43.6$\pm$0.8} & 55.8$\pm$0.6 & \textbf{34.9$\pm$0.7} & 43.1$\pm$0.6 & 62.0$\pm$0.7 \\
AutoPalette & 19.4$\pm$2.5 & 32.0$\pm$1.1 & 56.8$\pm$0.7 & 28.3$\pm$1.1 & 39.2$\pm$1.1 & 61.7$\pm$0.5 & 29.7$\pm$2.1 & 50.1$\pm$1.3 & 61.3$\pm$0.5 \\
\rowcolor{gray!14}
QuADD (Ours) & 21.4$\pm$0.7 & 29.7$\pm$0.7 & 50.3$\pm$0.8 & 29.8$\pm$1.0 & 39.6$\pm$0.7 & \textbf{63.5$\pm$0.5} & 30.7$\pm$1.2 & \textbf{51.8$\pm$0.5} & \textbf{66.6$\pm$0.4} \\
\bottomrule
\end{tabular}%
}
\caption{Accuracy comparison of different methods across AlexNet, VGG, and ResNet under varying IPC values on CIFAR-10. Best results in each column are highlighted in bold.}
\label{tab:cross_arch}
\end{table*}

\begin{table*}[t]
\centering
\renewcommand{\arraystretch}{1.2}
\begin{tabular}{l|ccc|ccc}
\toprule
\multirow{2}{*}{Method} 
  & \multicolumn{3}{c|}{DM} 
  & \multicolumn{3}{c}{TM} \\
\cmidrule(lr){2-4} \cmidrule(lr){5-7}
 & IPC1 & IPC10 & IPC50 & IPC1 & IPC10 & IPC50 \\
\midrule
\textit{Vanilla (unquantized)} & 26.0 $\pm$ 0.8 & 48.3 $\pm$ 0.6  & 63.0 $\pm$ 0.4 & 45.7 $\pm$ 0.8 & 65.2 $\pm$ 0.7 & 71.4 $\pm$ 0.2\\
\midrule
FreD        & \textbf{33.6$\pm$ 0.4}  & \textbf{48.1$\pm$ 0.8}  & 62.0 $\pm$ 0.6 & 43.1$\pm$ 0.3 & 57.3 $\pm$ 0.5 & 69.0 $\pm$ 0.2 \\
AutoPalette   & 12.0 $\pm$ 0.4 & 15.3 $\pm$ 0.2 & 48.5 $\pm$ 0.4 &  42.0 $\pm$1.1 & 61.2 $\pm$ 0.2 & 69.5 $\pm$ 0.2 \\
\rowcolor{gray!14}
QuADD (Ours) & 17.5 $\pm$ 0.5 & 45.1 $\pm$ 0.3 & \textbf{62.4$\pm$ 0.3}  & \textbf{44.7$\pm$ 0.5}  & \textbf{63.2$\pm$ 0.2}  & \textbf{70.5$\pm$ 0.2} \\
\bottomrule
\end{tabular}
\caption{Performance comparison across distillation methods on DM and TM datasets under varying IPC values on CIFAR-10. The unquantized baseline serves as a reference for all methods. Best results in each column are highlighted in bold.}
\label{tab:quadd_cross_dd}
\end{table*}

\subsection{Cross-Domain Evaluation: Tabular 3GPP Data}\label{Sec:exp_3gpp}
To assess QuADD’s generality beyond vision tasks, we evaluate it on a beam management problem inspired by 3GPP wireless systems.
The goal is to predict the optimal transmission beam (beam index) from a partially observed set of channel measurements representing user-side signal statistics such as received power.
These scalar, unstructured features serve as a rigorous benchmark to evaluate QuADD’s multimodal capabilities.

\medskip
\noindent \textbf{Problem Description.}
The task involves identifying the transmit–receive beam pair with the highest received power from a masked set of beam measurements (see Appendix \ref{sec:beam_meta} for details).
The received measurements are organized as a matrix of beam pairs, where masking prevents full observation of all entries.
The model takes this partially observed matrix as input and aims to interpolate missing values to predict the beam with the maximum reconstructed power, making the task analogous to a matrix completion problem.

\medskip
\noindent \textbf{DD Setup.}
The true data contains $3\times10^4$ measurements generated from a simulated mmWave link setup.
A three-layer CNN is trained on distilled datasets with $M$ samples per class and bit precision $B \in \{4, 8, 32\}$.
All methods are evaluated under a fixed storage budget ($M \times B \times D$ bits) to ensure fair comparison across full-precision, quantized-only, and quantization-aware distilled data.

\begin{figure}
    \centering    \includegraphics[width=1.0\linewidth]{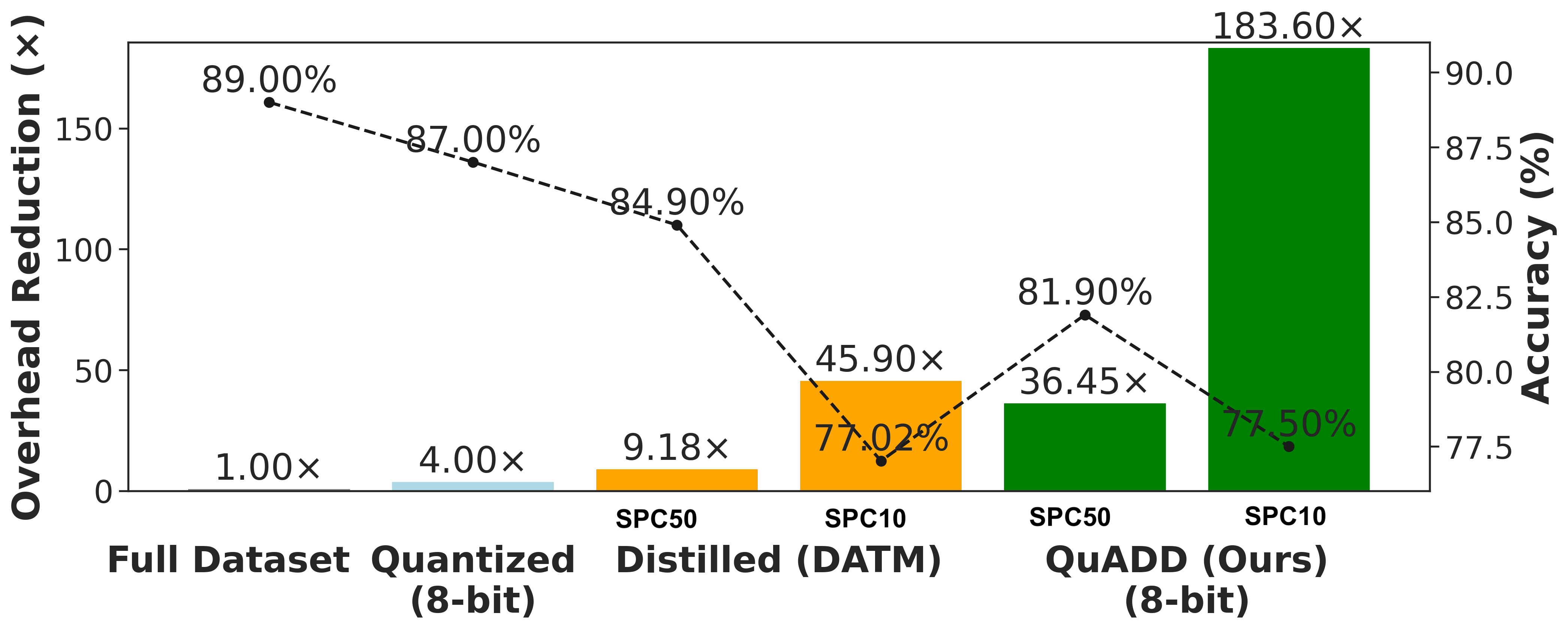}
    \caption{Overhead reduction and prediction accuracy on beam management problem. Higher is better.}
    \label{fig:beam1}
\end{figure}

\medskip
\noindent \textbf{Results.}
Figure~\ref{fig:beam1} shows the trade-off between compression and accuracy on the beam management task.
Bars indicate compression ratios, and the dashed line shows accuracy.
A model trained on the full dataset reaches 89\% accuracy.
Quantizing the full dataset (8-bit) yields $4\times$ reduction with accuracy 87\%, while unquantized DD gets 77\% accuracy at $46\times$ reduction. QuADD achieves 81.9\% at $36\times$ reduction, and 77.5\% accuracy at a substantial $183\times$ reduction.
\begin{figure}[t]
\centering
\includegraphics[width=0.81\columnwidth]{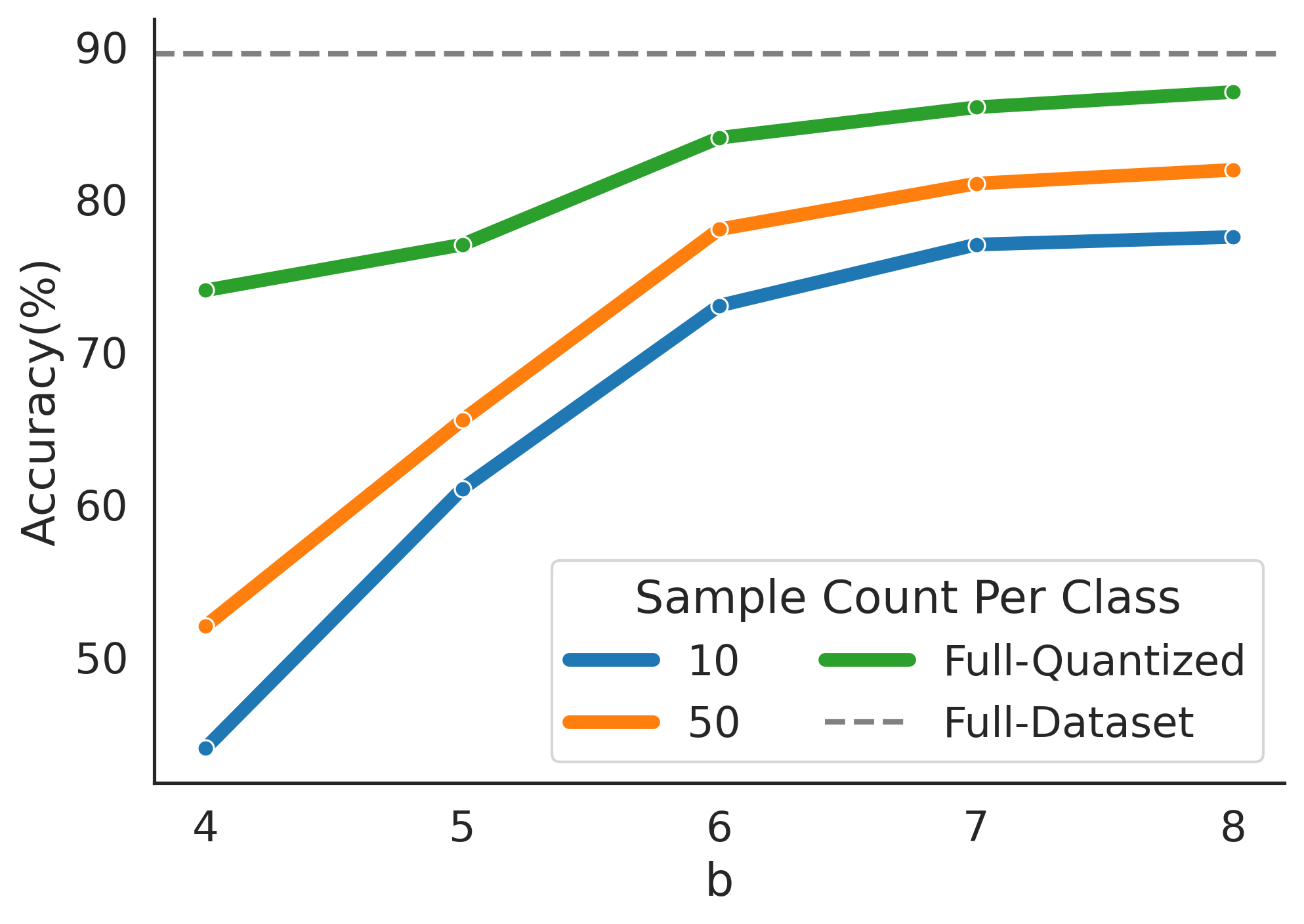}
\caption{
Accuracy–rate trade-off on the 3GPP beam management dataset for different sample counts (SPC = 10, 50, Full).}
\label{fig:beam_rate_distortion}
\end{figure}

\noindent
Figure~\ref{fig:beam_rate_distortion} illustrates QuADD’s rate–distortion trade-off on the beam management dataset at different SPCs (10, 50, and full).
Accuracy rises with bit precision across all scales but plateaus beyond 6 bits, indicating that moderate precision suffices to capture most task-relevant information.

\section{Conclusions}
\noindent
This work redefines DD through the lens of information efficiency, shifting the goal \textit{from fewer samples to fewer bits}.
Unlike prior approaches that minimize sample count, we show that jointly optimizing sample count and precision yields more compact yet equally informative distilled data.

\noindent
We introduced \textbf{Quantization-aware Dataset Distillation (QuADD)}, a framework that integrates quantization into the distillation loop.
By jointly adapting synthetic data and quantization parameters end-to-end, QuADD produces low-precision datasets that retain true data information.
Its adaptive non-uniform quantization layer further aligns quantization density with data, improving representational efficiency under stringent bit budget constraints.

\noindent
Experiments on image benchmarks and a 3GPP task show that QuADD delivers state-of-the-art performance under fixed bit budgets.
Notably, QuADD preserves near-baseline accuracy while compressing data by over $10\times$ on image datasets and more than $180\times$ on 3GPP data, highlighting its effectiveness and generality beyond visual domains.

\newpage
\bibliography{main_references}
\bibliographystyle{ieeenat_fullname}

\clearpage
\appendix
\setcounter{page}{1}
\maketitlesupplementary  

\section{Baseline: Uniform Quantization}
\label{sec:uniform}
\subsection{Overview}
Uniform quantization serves as a baseline implementation of the quantization layer. Given bit precision $b$ and clipping threshold $\alpha$, the uniform quantizer applies evenly spaced levels after clipping:

\begin{equation}
Q^u(x;\alpha,b)=\mathrm{round}\;\left(\frac{ \lfloor x,\alpha \rceil}{\Delta}\right)\Delta,
\quad 
\Delta=\frac{2\alpha}{2^{b}-1}.
\end{equation}

\noindent
Uniform quantization is widely used in deep image compression due to its simplicity and computational efficiency \cite{tsubota2023comprehensivecomparisonsuniformquantization}. However, because the rounding operator is nondifferentiable with zero gradients almost everywhere, several surrogate gradient techniques are used during training.

\noindent
Let $z = \lfloor x,\alpha \rceil$. The following formulations are the most common.

\noindent \textbf{1. Straight-Through Estimator (STE):}

\noindent Forward: $$Q_{\text{STE}}(z) = \mathrm{round}(z/\Delta)\Delta.$$

\noindent Backward: $$\frac{\partial Q_{\text{STE}}}{\partial z} \approx 1$$.

\noindent
This treats quantization as the identity function during backpropagation. STE is simple but introduces an entropy estimation gap because it does not model quantization noise during training.

\noindent
\textbf{2. Additive Uniform Noise (AUN):}

\noindent
Forward:
$$Q_{\text{AUN}}(z) = z + u, \qquad u \sim \mathcal{U}\!\left[-\tfrac{\Delta}{2}, \tfrac{\Delta}{2}\right].$$

\noindent
Backward: Same as STE (gradient $\approx 1$).

\noindent AUN provides a continuous relaxation of rounding and is widely used in compression models, but it introduces a discrete gap since no actual rounding occurs during training. We do not use this approach in our experiments. 

\noindent \textbf{3. Soft / Differentiable Rounding:}
Soft rounding smooths the hard rounding transition using a continuous $k$-shaped function, reducing gradient mismatch and improving stability \cite{gong2019differentiablesoftquantizationbridging, mentzer2024finite}:
$$Q_{\text{soft}}(z) \approx \mathrm{softround}_k(z)$$
where the backward pass is governed by the derivative of the underlying hyperbolic functions (e.g., tanh). This reduces the gradient mismatch inherent to STE by smoothing the rounding boundary.

\noindent
\textbf{FSQ-Style Symmetric Uniform Quantizer} In our experiments, we adopt a modified FSQ scalar quantizer \cite{parker2025scaling} designed to maintain symmetry around the origin for any number of quantization levels $L = 2^b$. 

\noindent Forward: For a scalar input $x$, the forward quantization operator is:
\begin{equation}
\label{eq:fsq_sym}
      Q_L(x)
= 
\frac{2}{L-1}
\left\lfloor 
\frac{(L - 1)\tanh(x) + 1}{2}
+\frac{1}{2}
\right\rfloor
 - 1. 
\end{equation}
This maps $\tanh(x)\in(-1,1)$ onto a set of $L$ uniformly spaced levels in $[-1,1]$.

\noindent
Backward: A hybrid training strategy combining additive noise and STE backpropagation is used: 

\begin{itemize}
    \item Noise-based approximation (50\% of training steps): 
    $$
    Q_L(x) \approx \tanh(x) 
+ \frac{U\{-1,1\}}{L - 1},
    $$ where $U\{-1,1\}$ samples $-1$ or $+1$ uniformly.

    \item Straight-Through Estimator (remaining 50\%): In the other half of training steps, we use the hard quantizer from Eq. \ref{eq:fsq_sym} in the forward pass and the STE in the backward pass. 
\end{itemize}

\subsection{Benchmark Comparision:}
\label{sec:benchmark_comp}
\begin{figure*}
    \centering
\includegraphics[width=0.8\linewidth]{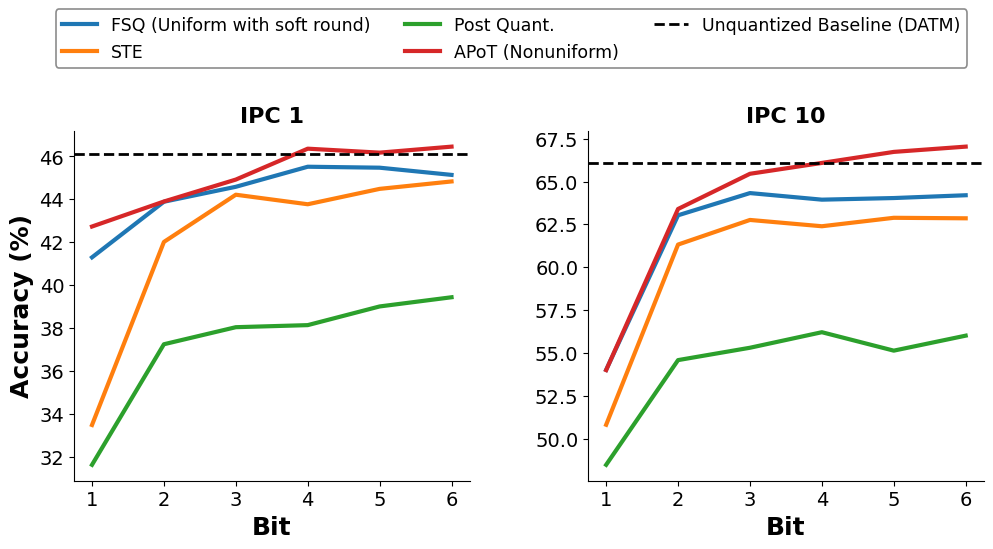}
    \caption{Benchmark comparison between uniform quantization variants and non-uniform (APoT) quantization on CIFAR-10 for IPC\,1 and IPC\,10. 
    Each curve evaluates synthetic data quantized from 1 to 6 bits per channel (resulting in $b\times 3$ bits per RGB pixel). 
    The unquantized baseline (DATM) is stored at standard 32-bit precision. }
    \label{fig:uni_vs_nonuni}
\end{figure*}

\noindent
Figure \ref{fig:uni_vs_nonuni} compares several uniform-quantization baselines against our non-uniform APoT quantizer on CIFAR-10 for IPC 1 and IPC 10 across bit-widths from 1 to 6. Across both IPC settings, all uniform methods improve as bit-precision increases, but their behaviors differ substantially. Post-training quantization performs worst, particularly at low precision, highlighting the mismatch created when synthetic data are optimized in full precision and later quantized. STE improves over post-quantization but remains sensitive to rounding noise. The FSQ-based differentiable uniform quantizer achieves consistently higher accuracy than other uniform variants, confirming the value of soft rounding and hybrid training for stabilizing optimization.

\noindent
APoT (non-uniform), however, achieves the strongest results overall—especially at the most aggressive precision levels—reflecting its ability to allocate more resolution near high-density regions of the synthetic data. Notably, APoT reaches or surpasses the unquantized DATM baseline with only 3–4 bits per channel, whereas uniform methods require higher precision to approach similar performance.

\section{Nonuniform Quantization }
\label{sec:apot_setting}

\paragraph{Normalization:}
APoT was originally designed for model quantization, where weight tensors are approximately Gaussian. Accordingly, \cite{Li2020Additive} normalizes weights to zero mean and unit variance before each clipping–projection step to stabilize training. However, this assumption does not directly transfer to dataset distillation, where the data distribution depends on the task and modality rather than following a near-Gaussian weight prior.

\noindent
In image classification, synthetic images retain channel-wise structure, so we apply per-channel batch normalization before quantization to standardize the activation range and improve stability. For wireless tabular data, feature distributions are heterogeneous and often non-stationary, making unified normalization counterproductive; in such cases, we skip normalization.

\paragraph{Initialization of the Clipping Threshold $\alpha$:}
The clipping threshold $\alpha$ determines the dynamic range of the quantizer and strongly affects quantization error. Too large $\alpha$ widens the quantization grid and increases projection error in the high-density central region, while too small 
$\alpha$ causes excessive clipping of informative values. Learning 
$\alpha$ directly is unstable, as noted in \cite{Li2020Additive}, because the data distribution shifts constantly throughout training.

\noindent
To ensure faster and more reliable convergence, we initialize $\alpha$ using a robust percentile-based range:
\begin{equation}
    \alpha_0=\big |\mathrm{P}_{99}(x)-\mathrm{P}_{1}(x)\big|,
\end{equation}

\noindent which removes extreme outliers while retaining most of the useful signal. 

\noindent
For settings that apply DCT-based transformations to the data, as in \cite{shin2024}, we follow a two-band frequency strategy: low-frequency components and high-frequency components each receive their own learnable clipping threshold. We initialize these as: 
\begin{equation}
    \begin{aligned}
    \alpha^{\text{low}}_0&=\big|\mathrm{P}_{99}(x_{\text{low}})-\mathrm{P}_{1}(x_{\text{low}})\big|, \\
\alpha^{\text{high}}_0 &=\big|\mathrm{P}_{99}(x_{\text{high}})-\mathrm{P}_{1}(x_{\text{high}})\big|
    \end{aligned}
\end{equation}

\noindent
reflecting the different dynamic ranges across frequency bands. 

\section{Experiments Details}
\label{sec:hyperparam}

\paragraph{Datasets:}
We conduct experiments across both image and tabular modalities:
\begin{itemize}
    \item CIFAR-10: A 10-class image dataset containing 50,000 training and 10,000 test RGB images of size $32\times32$.  Classes include airplane, automobile, bird, cat, deer, dog, frog, horse, ship, and truck.
    \item CIFAR-100: Similar in structure to CIFAR-10 but with 100 classes and 600 images per class (500 train, 100 test). Image resolution is also in $32\times32$.
    \item ImageNette Subset: 10-class high-resolution subsets of ImageNet 
    ,containing $128\times128$ RGB images. 
   \item 3GPP Beam Management (Wireless Tabular): A tabular dataset for wireless signal prediction, consisting of 29,784 samples split into 80\% training and 20\% testing. Each sample has 64 features and one of 64 beam classes. Unlike the image datasets, the class distribution is highly imbalanced, with class sizes ranging from 14 to 284 samples.
\end{itemize}

\paragraph{Networks}

For low-resolution datasets (CIFAR-10/100), we adopt the standard ConvNetD3 architecture used in prior distillation works. The network contains 3 convolutional blocks, each with 128 filters of size $3 \times 3$,  followed by an  instance normalization layer, a ReLU, and an average pooling layer with $2 \times 2$ kernel and stride 2. For ImageNet subsets, we follow prior practice and use ConvNet-D5, a deeper 5-layer variant with the same block structure.

\noindent
For cross-architecture evaluation, we train distilled datasets on three unseen networks: VGG11\cite{simonyan2014very}, AlexNet\cite{krizhevsky2009learning}, and ResNet18\cite{he2016deep}. All architectures follow their standard implementations.

\noindent
For the palette network used in \cite{colornips2024}, we follow their architecture and employ a single $1\times1$ convolutional layer. The palette network is first warmed up for 2 epochs before being jointly optimized within the DD pipeline.

\begin{table*}[t]
\centering
\small
\caption{Hyperparameters for all methods based on Trajectory Matching (TM) framework.}
\label{tab:tm_params}
\resizebox{\textwidth}{!}{%
\begin{tabular}{lcccccccccc}
\toprule
Dataset & IPC & Synth. BS & Synth. steps & Expert ep. & Max start ep. &
Quant. LR & Img. LR & Step LR & Teacher LR & ZCA \\
\midrule
CIFAR-10
 & 1  & 200  & 80 & 2 & 15 & 0.1 & 500  & $10^{-7}$ & $10^{-2}$ & True \\
 & 10 & 400  & 35 & 2 & 40 & 0.1 & 1000 & $10^{-5}$ & $10^{-2}$ & True \\
 & 50 & 200 & 60 & 2 & 40 & 0.1 & 500  & $10^{-5}$ & $10^{-2}$ & True \\
\midrule
CIFAR-100
 & 1  & 200  & 60 & 2 & 25   & 0.1 & 1000 & $10^{-5}$ & $10^{-2}$ & True \\
 & 10 & 500 & 50 & 2 & 70 & 0.1 & 1000 & $10^{-5}$ & $10^{-2}$ & True \\
 & 50 & 800 & 80 & 2 & 70 & 0.1 & 1000 & $10^{-5}$ & $10^{-2}$ & True \\
\midrule
ImageNette
 & 10 & 60 & 40 & 2 & 20 & 0.1 & 1000 & $10^{-5}$ & $10^{-2}$ & False \\
\bottomrule
\end{tabular}
}
\end{table*}

\begin{table*}[t]
\centering
\caption{Hyperparameters for our method based on the Distribution Matching (DM) framework.}
\label{tab:dm_hparams}
\begin{tabular}{lccccc}
\toprule
Dataset & IPC & Synthetic batch size & LR (Quantizer) & LR (Synthetic Image) & ZCA \\
\midrule
 CIFAR-10& 1  & --  & 0.1 & 1  & True \\
 & 10 & -- & 0.1 & 1  & True \\
 & 50 & -- & 0.1 & 10 & True \\
\midrule
 CIFAR-100 & 1  & -- & 0.1 & 1  & True \\
 & 10 & --  & 0.1 & 1  & True \\
 & 50 & 50  & 0.1 & 10 & True \\
\bottomrule
\end{tabular}
\end{table*}

\paragraph{Implementation Details}

For APoT quantization layer, the module consists of a learnable clipping threshold $\alpha$, a
$b$-bit precision parameter, and an optional batch-normalization step applied before clipping when enabled.

\noindent
QuADD is trained using Trajectory Matching (TM) by default, though it is fully compatible with other DD objectives such as Distribution Matching (DM). The hyperparameter settings for the non-uniform quantization case, as well as the configurations used for benchmark methods (AutoPalette, FReD, and DATM), are provided in Tables~\ref{tab:tm_params} and~\ref{tab:dm_hparams} for each dataset and framework.

\noindent
For FReD \cite{shin2024}, we used the hyperparameter settings reported in the paper.

\noindent
We observe that QuADD is generally robust across hyperparameters, though several settings still require careful tuning. Many of these align with prior DD work—including the number of synthetic steps, the maximum start epoch, and the synthetic batch size.

\noindent
QuADD’s quantization layer itself introduces only a small number of hyperparameters, and we find that it requires minimal tuning in practice:

\begin{itemize}
    \item Uniform quantization: choice of companding function (e.g., \texttt{tanh}, Laplace CDF) or an optional linear transform to reshape the data distribution before quantization.
    \item Non-uniform quantization: number of learnable clipping thresholds~$\alpha$, which is domain-specific. For images, we use either a single $\alpha$ shared across RGB channels or one per channel; for tabular data, $\alpha$ may be per feature or shared globally. Additional $\alpha$ parameters yield only marginal gains (typically $\leq 1\%$), so we adopt the minimal setting unless noted otherwise.
    \item Batch normalization: applied before quantization. For uniform quantization, we find it beneficial primarily when the data is not preprocessed with ZCA. For nonuniform quantization, it is applied by default in the image case.

\end{itemize}

\noindent
All experiments use the SGD optimizer with standard DD training schedules.
Unless otherwise specified:

\begin{itemize}
    \item CIFAR experiments use $32\times 32$ synthetic images at 2–6 bits per sub-pixel.
    \item ImageNet subset experiments use $128\times 128$ synthetic images at 3–5 bits.
    \item Wireless experiments use 4-8-bit quantization.
\end{itemize}

\noindent
Training was performed on 4 $\times$ NVIDIA V100 (32GB) or 2$\times$ NVIDIA H100 (80GB) GPUs.

\section{QuADD Implementation for DATM}
We integrate our proposed Quantization-aware Dataset Distillation (QuADD) framework with the Difficulty-Aligned Trajectory Matching (DATM) method~\cite{guo2024lossless}, forming a quantization-aware variant referred to as \textbf{QuADD-DATM}. This integration enables simultaneous optimization of synthetic data, model parameters, and quantizer precision under fixed bit budgets.

\paragraph{DATM Objective.}
DATM optimizes a synthetic dataset $\mathcal{S}$ such that model parameters $\theta$ trained on $\mathcal{S}$ follow a trajectory that closely matches the training trajectory on the real dataset $\mathcal{T}$. The DATM loss for a model parameterized by $\theta_t$ at iteration $t$ is expressed as
\begin{equation}
    \begin{aligned}
    \mathcal{L}_{\text{DATM}}(\mathcal{S}; \theta_t)
    &= \sum_{t=1}^{T} 
    \| \theta_{t+1}^{\mathcal{T}} - \theta_{t+1}^{\mathcal{S}} \|_2^2,
    \\
    \theta_{t+1}^{\mathcal{S}} &= \theta_t - \eta \nabla_{\theta_t} \mathcal{L}_{\text{CE}}(\mathcal{S}, \theta_t),
    \end{aligned}
    \label{eq:datm_loss}
\end{equation}
where $\eta$ is the learning rate, and $\mathcal{L}_{\text{CE}}$ denotes the cross-entropy loss. 

\paragraph{Quantization-Aware Formulation.}
In QuADD-DATM, we introduce a differentiable quantization layer $Q(\cdot; \alpha, b)$ into the synthetic data path, such that the quantized synthetic dataset is
\begin{equation}
    \mathcal{S}_q = Q(\mathcal{S}; \alpha, b),
\end{equation}
where $\alpha$ is the clipping threshold and $b$ is the bit precision. The quantizer used is the APoT quantization layer, with the implementation of the forward and backward pass detailed in Sec. \ref{Sec:adaptive_quant}.

\noindent 
The QuADD-DATM objective is therefore modified as
\begin{equation}
    \begin{aligned}
    \mathcal{L}_{\text{QuADD-DATM}} &= 
    \sum_{t=1}^{T} 
    \| \theta_{t+1}^{\mathcal{T}} - \theta_{t+1}^{\mathcal{S}_q} \|_2^2,
    \\
    \theta_{t+1}^{\mathcal{S}_q} &= \theta_t - \eta \nabla_{\theta_t} \mathcal{L}_{\text{CE}}(\mathcal{S}_q, \theta_t).
    \end{aligned}
    \label{eq:quadd_datm_loss}
\end{equation}

\paragraph{Algorithm Overview.}
Algorithm~\ref{alg:quadd_datm} summarizes the QuADD-DATM procedure. At each iteration, real and synthetic mini-batches are sampled, quantized through $Q(\cdot)$, and used to compute the trajectory-matching loss. Both $\mathcal{S}$ and the quantizer parameters are updated jointly, enabling adaptation to precision-induced distortion.

\begin{algorithm}[t]
\caption{QuADD Implementation for DATM with APoT}
\label{alg:quadd_datm}
\KwInput{Real dataset $\mathcal{T}$, synthetic dataset $\mathcal{S}$, APoT quantizer $Q_{\text{APoT}}(\cdot;\alpha,b,k,n,\gamma)$, trajectory mapping $\phi(\cdot;\theta)$, learning rates $\eta_S,\eta_\alpha,\eta_\gamma,\eta_\theta$}
\KwOutput{Quantized distilled dataset $\mathcal{S}^{q^{*}} = Q_{\text{APoT}}^{*}(\mathcal{S}^{*})$}
Initialize $\mathcal{S}$ and APoT parameters $(\alpha,\gamma)$ as in Sec.~\ref{Sec:adaptive_quant}; fix $(b,k,n)$\;
\For{each distillation iteration $t=1,\dots,T$}{
    \tcp{Mini-batch sampling}
    Sample $\mathcal{B}_{\mathcal{T}}\!\sim\!\mathcal{T}$ and $\mathcal{B}_{\mathcal{S}}\!\sim\!\mathcal{S}$\;
    \tcp{Quantize with APoT (Sec.~\ref{Sec:adaptive_quant})}
    $\mathcal{S}^q \leftarrow Q_{\text{APoT}}(\mathcal{B}_{\mathcal{S}})$\;
    \tcp{DATM inner loop (one step, referenced)}
    Compute per-sample CE losses and logits on $\mathcal{B}_{\mathcal{T}}$ and $\mathcal{S}^q$ with $\theta_t$ in \eqref{eq:datm_loss}\;
    \tcp{Trajectory alignment on quantized data}
    Evaluate $\mathcal{L}_{\text{QuADD-DATM}}$ using \eqref{eq:quadd_datm_loss}\;
    \tcp{Backprop through APoT quantizer}
    Backpropagate using \eqref{eq:grad_chain} \;
    \tcp{Parameter updates (keep $(b,k,n)$ fixed)}
    $\mathcal{S} \leftarrow \mathcal{S} - \eta_S \frac{\partial \mathcal{L}}{\partial \mathcal{S}}$;\quad
    $\alpha \leftarrow \alpha - \eta_\alpha \frac{\partial \mathcal{L}}{\partial \alpha}$;\quad
    $\gamma \leftarrow \gamma - \eta_\gamma \frac{\partial \mathcal{L}}{\partial \gamma}$\;
}
\end{algorithm}


\begin{figure*}
    \centering
    \includegraphics[width=1\linewidth]{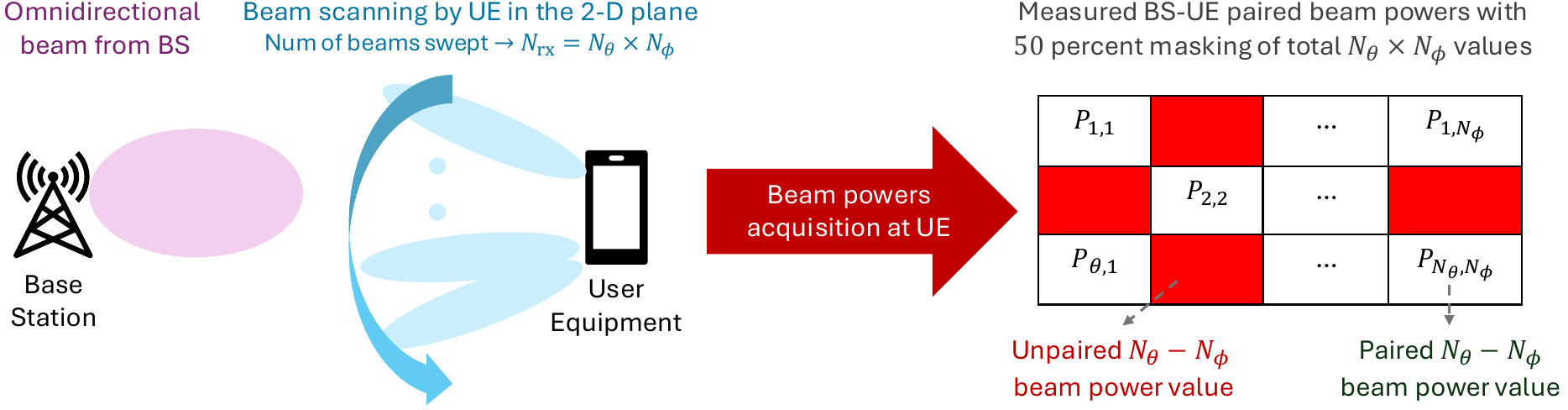}
    \caption{\textbf{Illustration of the 3GPP beam management problem under one-sided beam scanning.}
    The base station (BS) transmits an omnidirectional or fixed reference beam, while the user equipment (UE) performs directional scanning across azimuth ($\theta$) and elevation ($\phi$) angles. 
    The resulting $N_{\theta} \times N_{\phi}$ grid of received beam powers represents the UE’s spatial response pattern. 
    Only a subset of these directions—here, 50\%—is measured, simulating limited probing due to time or energy constraints. 
    The masked and observed beam powers form a tabular dataset used as input to the AI/ML model, with missing entries representing unmeasured spatial directions.}
    \label{fig:3gpp_bm_data_acq}
\end{figure*}
\section{Beam-management problem for 3GPP wireless communication systems}
\label{sec:beam_meta}
\subsection{Overview}
\paragraph{Role of AI/ML in 3GPP wireless systems}
Artificial Intelligence and Machine Learning (AI/ML) have become integral components of modern wireless systems, especially within the 3GPP standardization framework. Starting from Release 17 and continuing through Releases 18–20, AI/ML methods are being systematically explored for radio access network (RAN) optimization, air-interface design, and system-level management. The goal is to improve network adaptability and efficiency under diverse environmental conditions and operational constraints. Within 3GPP, AI/ML techniques are primarily leveraged to:
\begin{itemize}
    \item Enhance network automation via data-driven models that can dynamically learn from radio measurements and user contexts.
    \item Optimize signal processing tasks such as channel estimation, link adaptation, and beam selection.
    \item Reduce signaling and feedback overhead, by learning compact representations of high-dimensional wireless environments.
    \item Enable cross-layer intelligence, where data from physical, MAC, and higher layers are fused for predictive or prescriptive control.
\end{itemize}

\paragraph{Beam Management Problem in 3GPP Wireless Systems}
Beam management is a cornerstone of millimeter-wave (mmWave) and sub-THz communications, where highly directional transmissions are necessary to overcome path loss and fading\footnote{D. Tse and P. Viswanath, \textit{Fundamentals of Wireless Communication}, Cambridge University Press, 2005.}\footnote{H. L. Van Trees, \textit{Optimum Array Processing: Part IV of Detection, Estimation, and Modulation Theory}, John Wiley \& Sons, 2002.}\footnote{E. Bj{\"o}rnson, J. Hoydis, and L. Sanguinetti, ``Massive MIMO networks: Spectral, energy, and hardware efficiency,'' \textit{Foundations and Trends in Signal Processing}, vol. 11, no. 3--4, pp. 154--655, 2017.}. In the 3GPP framework, beam management encompasses the procedures for beam sweeping, measurement, reporting, and selection between the base station (BS) and user equipment (UE) \footnote{3rd Generation Partnership Project (3GPP), ``Study on Artificial Intelligence (AI)/Machine Learning (ML) for NR air interface,'' Technical Report (TR) 38.843, V18.0.0, Dec. 2023. [Online]. Available: \url{https://www.3gpp.org/ftp/Specs/archive/38_series/38.843/}}\footnote{3rd Generation Partnership Project (3GPP), ``NR; Medium Access Control (MAC) protocol specification,'' Technical Specification (TS) 38.321, V17.7.0, Dec. 2023. [Online]. Available: \url{https://www.3gpp.org/ftp/Specs/archive/38_series/38.321/}}\footnote{3rd Generation Partnership Project (3GPP), ``Study on New Radio Access Technology; Physical Layer Aspects,'' Technical Report (TR) 38.802, V14.2.0, Sept. 2017. [Online]. Available: \url{https://www.3gpp.org/ftp/Specs/archive/38_series/38.802/}}.


The goal of beam management is to identify the optimal transmit–receive (Tx–Rx) beam pair that maximizes the received power or signal-to-noise ratio (SNR) at the UE. However, this search space grows quadratically with the number of Tx and Rx beams—e.g., with $N_{\mathrm{tx}}$ transmit beams and $N_{\mathrm{rx}}$ receive beams, the system must evaluate $N_{\mathrm{tx}}\times N_{\mathrm{rx}}$ possible beam pairs. Measuring all combinations is often infeasible due to time and energy constraints. The AI/ML formulation seeks to predict the best beam pair from a partially observed subset of measurements—thereby reducing beam sweeping overhead while maintaining near-optimal link quality.

The BM framework is portrayed in Fig. \ref{fig:3gpp_bm_data_acq}. Although the general beam management problem in 3GPP systems involves identifying the optimal transmit–receive (Tx–Rx) beam pair, we illustrate here a simplified one-sided scanning scenario, where the base station (BS) transmits an omnidirectional or fixed reference beam, and the user equipment (UE) performs directional scanning in both azimuth and elevation planes. The UE thus collects a set of received beam power measurements corresponding to $N_{\theta}\times N_{\phi}$, representing its local angular response to the BS transmission. Only a subset of these measurements—here, 50 \% of the total entries—are available due to masking, emulating practical limitations in beam probing caused by time and energy constraints. The measured powers $P_{i,j}$ are organized in a 2-D tabular form indexed by azimuth $(\theta)$ and elevation $(\phi)$, where missing entries denote unobserved directions. 

\begin{figure}
    \centering
    \includegraphics[width=1\linewidth]{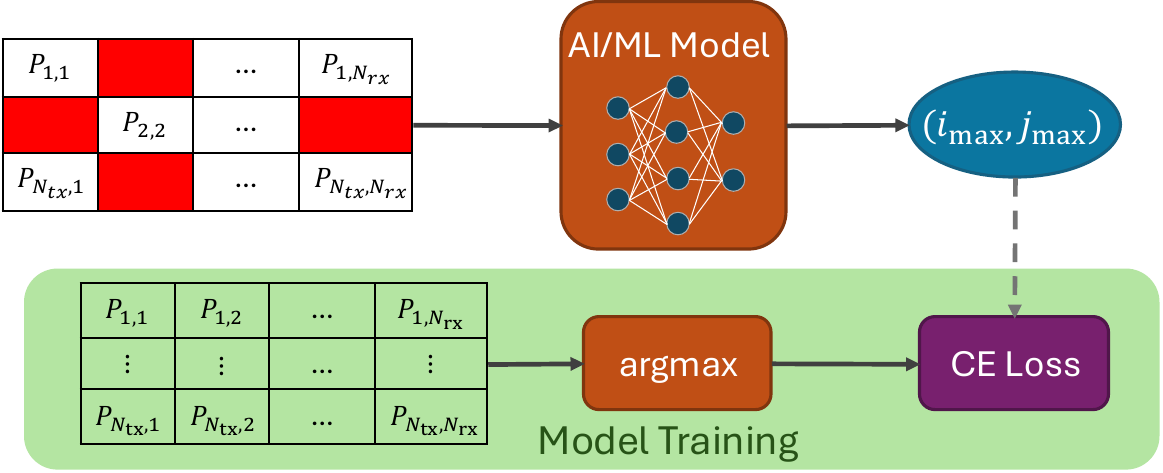}
    \caption{\textbf{AI/ML-based inference for optimal beam prediction in the 3GPP beam management problem.}
    The masked beam-power matrix from Fig.~\ref{fig:3gpp_bm_data_acq} is processed by an AI/ML model trained to nonlinearly interpolate missing measurements and predict the beam direction corresponding to the maximum reconstructed power. 
    The model outputs the estimated index $(i_{\text{max}}, j_{\text{max}})$ of the optimal beam, trained using a cross-entropy loss over all possible beam indices.}
    \label{fig:3gpp_bm_ml_inf}
\end{figure}
An AI/ML model is used for the extraction of the best beam pairs from the observed measurements, as shown in Fig. \ref{fig:3gpp_bm_ml_inf}. This setup can be viewed as a matrix completion plus classification task, where the model must infer latent spatial correlations between adjacent beams to accurately estimate the unobserved entries. Such correlations often capture the angular continuity of propagation paths, reflecting realistic 3GPP channel models.
\begin{itemize}
    \item The AI/ML model receives the masked beam-power matrix as input.
    \item The network learns to nonlinearly interpolate the missing values and predict the beam index $(i_{\mathrm{max}},j_{\mathrm{max}})$ corresponding to the maximum received power.
    \item The model is trained using cross-entropy loss (CE Loss) to classify the correct optimal beam among all candidates. The training data consists of complete tabular measurements with all $N_{\mathrm{tx}}\times N_{\mathrm{rx}}$ possible beam pairs.
\end{itemize}

\paragraph{Dataset distillation for beam management in 3GPP systems}
In wireless communication systems, dataset distillation (DD) serves a particularly vital role owing to the distributed nature of data acquisition and model training across multiple base station (BS) and user equipment (UE) nodes. Each node observes channel conditions unique to its spatial, temporal, and environmental context, leading to inherently non-identically distributed (non-IID) data. Transferring large-scale, full-precision datasets from multiple nodes to a centralized learning server—or between cooperating edge nodes—imposes substantial communication and storage overhead under constrained fronthaul and backhaul bandwidths.
DD mitigates these challenges by synthesizing compact yet information-preserving datasets that capture the key statistical and structural features of local measurements. When shared across nodes, these distilled datasets significantly reduce the bit-level transfer cost while maintaining the fidelity required for downstream model training. Hence, in wireless systems, DD extends beyond mere data compression—it functions as a foundational enabler of efficient multi-node learning and collaboration, aligning with emerging 3GPP initiatives on AI/ML-driven air-interface optimization and distributed RAN intelligence.

\noindent The beam management dataset serves as a non-visual tabular benchmark for validating the generality of the proposed Quantization-aware Dataset Distillation (QuADD) framework.
While visual datasets like CIFAR or ImageNette test perceptual fidelity, the 3GPP beam management data evaluates QuADD’s performance in a domain characterized by structured sparsity and physical constraints. The distilled and quantized datasets aim to preserve predictive power while drastically reducing bit-level storage—reflecting practical needs in bandwidth-limited network environments.

\subsection{Ablation Study: Imbalanced Dataset}

\begin{figure}
    \centering
    \includegraphics[width=1.0\linewidth]{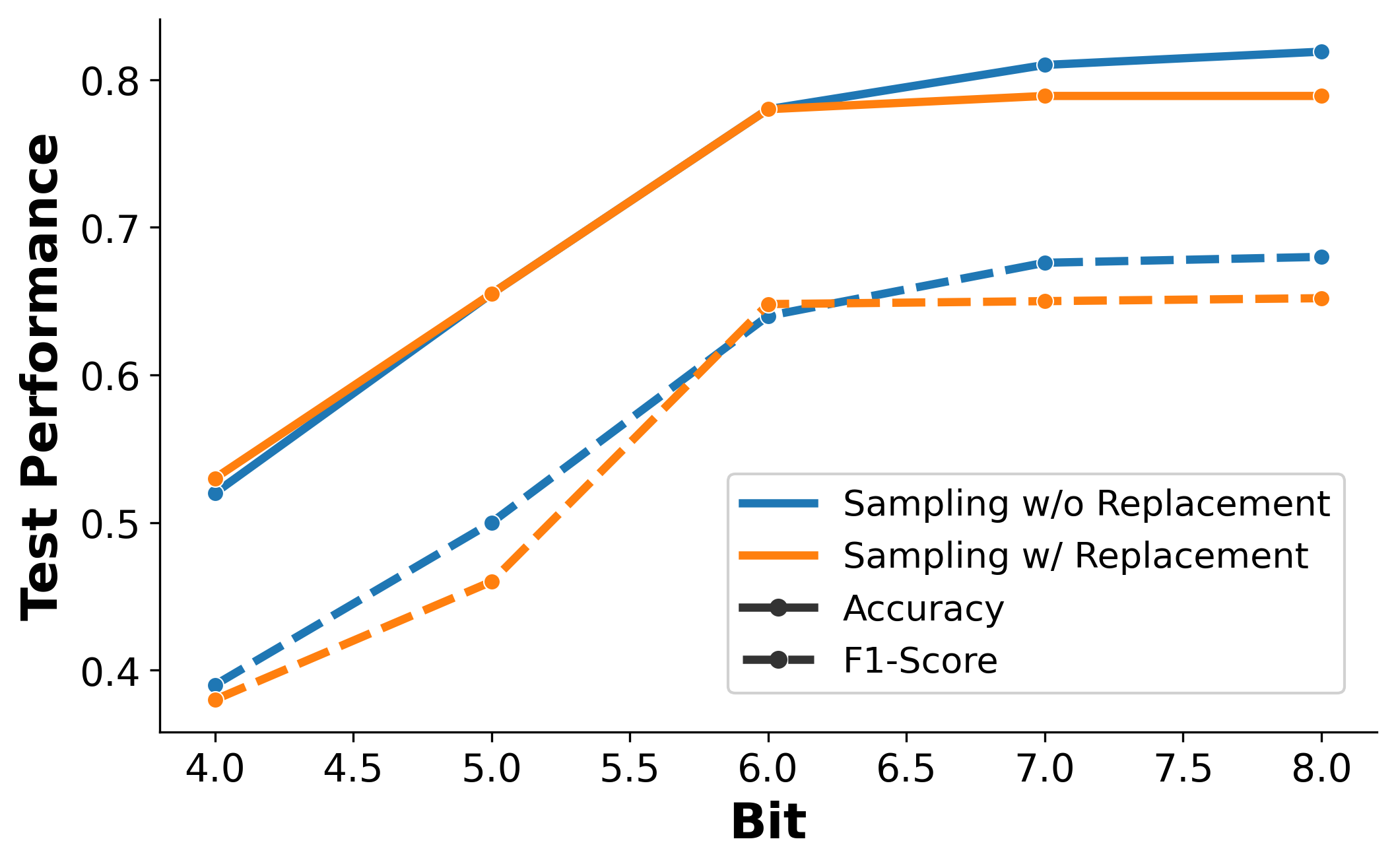}
    \caption{ Effect of sampling strategy on the imbalanced 3GPP wireless dataset at IPC\,50. 
    For classes with fewer than 50 real samples, we initialize $\mathcal{S}$ either by 
    (i) using all available samples (sampling \emph{without} replacement), or 
    (ii) duplicating samples to reach 50 via sampling \emph{with} replacement. }
    \label{fig:imbalance_sampling}
\end{figure}

Figure~\ref{fig:imbalance_sampling} presents an interesting case study on the imbalanced 3GPP wireless dataset—reflective of many real-world scenarios where class frequencies vary widely. At IPC,50, both sampling strategies improve with larger bit budgets, but sampling without replacement yields slightly higher Accuracy and F1-score, especially at 6–8 bits. One intuition is that sampling with replacement forces minority classes to be duplicated, which can overemphasize a few low-variation or noisy examples and reduce the effective diversity of the initialization. In contrast, using each available sample once preserves the natural variability of rare classes and avoids reinforcing redundant gradients, leading to more stable distillation under imbalance at higher fidelity.

\end{document}